\documentclass[runningheads]{llncs}
\pdfoutput=1
\usepackage{graphicx}
\usepackage{tikz}
\usetikzlibrary{backgrounds}
\usepackage[final]{microtype}
\usepackage{amssymb}
\usepackage{amsmath}
\usepackage{amsfonts}
\usepackage{subcaption}
\usepackage{siunitx}
\renewcommand{\epsilon}{\varepsilon}

\begin{document}
\title{Metrics for Learning in Topological Persistence}

\author{
Henri Riihim\"{a}ki\inst{1} 
\and
Jos\'e Lic\'on-Sal\'aiz\inst{2}\orcidID{0000-0002-8733-2256}
}

\authorrunning{H. Riihim\"{a}ki et al.}
%
\institute{Tampere University, Korkeakoulunkatu 7, 33720 Tampere, Finland \and
    Mathematical Institute, University of Cologne, Weyertal 86-90, 50931 Cologne, Germany\\ 
\email{henri.riihimaki@tuni.fi,licon@math.uni-koeln.de}}

\maketitle              
\begin{abstract}
Persistent homology analysis provides means to capture the connectivity structure of data sets in various dimensions. On the mathematical level, by defining a metric between the objects that persistence attaches to data sets, we can stabilize invariants characterizing these objects. We outline how so called contour functions induce relevant metrics for stabilizing the rank invariant. On the practical level, the stable ranks are used as fingerprints for data. Different choices of contour lead to different stable ranks and the topological learning is then the question of finding the optimal contour. We outline our analysis pipeline and show how it can enhance classification of physical activities data. As our main application we study how stable ranks and contours provide robust descriptors of spatial patterns of atmospheric cloud fields.

\end{abstract}
%
%
%
\section{Persistence pipeline}\label{sec:persistence_pipeline}
\subsection{Modelling data spaces}
Topological data analysis (TDA) and particularly its subfield persistent homology, or persistence, aim at quantifying the global connectivity structure of data sets \cite{CarlssonPointCloud,Oudot,Roadmap}. Given a set of data points it is often possible to endow it with some reasonable notion of relation between points, e.g. distance measure or correlation. Study of the connectivity is facilitated by first combining points into larger entities called simplices. A \(k\)-simplex is a declared subset of \(k+1\) related points from the data set. Collection of simplices makes up a simplicial complex \(C\), namely it is a collection of certain subsets of the data. Requirements are that if \(\sigma\) is a simplex in \(C\) then any subset of \(\sigma\) is also a simplex in \(C\) and that the intersection of two simplices is a simplex or the empty set. Above we have described an abstract simplicial complex. Simplices can always be realized geometrically in some \(\mathbf{R}^n\) as convex hulls of their vertices: 0-simplices as points, 1-simplices as line segments, 2-simplices as filled triangles, 3-simplices as filled tetrahedra etc. 

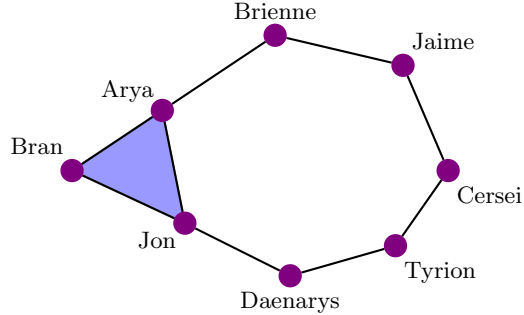
\begin{figure}[h]
	\centering
	\begin{tikzpicture}
	\tikzstyle{point}=[circle,thick,draw=violet,fill=violet,inner sep=0pt,minimum width=8pt,minimum height=8pt]
	
	\node[point](bran) at (0,0) {};
	\node[point](arya) at (1.2,0.8) {};
	\node[point](jon) at (1.5,-0.7) {};	
	\node[point](brienne) at (2.7,1.8) {};	
	\node[point](jaime) at (4.4,1.4) {};
	\node[point](daenarys) at (2.9,-1.4) {};
	\node[point](tyrion) at (4.3,-1) {};
	\node[point](cersei) at (5,0) {};	
	
	\node[above left] at (1.2,0.8) {Arya};
	\node[above left] at (0,0.1) {Bran};
	\node[below left] at (1.5,-0.7) {Jon};
    \node[above] at (2.7,1.9) {Brienne};
    \node[above right] at (4.4,1.5) {Jaime};
    \node[below] at (2.9,-1.5) {Daenarys};
    \node[below right] at (4.3,-1.1) {Tyrion};
    \node[below right] at (5,-0.1) {Cersei};
    
    \draw[thick,black] (arya) to (brienne);
    \draw[thick,black] (arya) to (bran);
    \draw[thick,black] (arya) to (jon);
    \draw[thick,black] (bran) to (jon);
    \draw[thick,black] (jon) to (daenarys);
    \draw[thick,black] (tyrion) to (daenarys);
    \draw[thick,black] (tyrion) to (cersei);
    \draw[thick,black] (cersei) to (jaime);
    \draw[thick,black] (brienne) to (jaime);
    
    \begin{scope}[on background layer]
    \filldraw[draw=black,thick,fill=blue,opacity=0.4] (1.2,0.8) -- (0,0) -- (1.5,-0.7) -- (1.2,0.8);
    \end{scope}    
	
	\end{tikzpicture}
	\caption{Simplicial model of social relations. To model relations between \(k+1\) points it is reasonable to use \(k\)-simplex for the purpose. Here the relations between \{Arya, Bran, Jon\} is depicted by the 2-simplex represented by the purple triangle. From the point of view of TDA the prominent feature of this data is the single loop structure, whereas a graph would see two loops (the closed path (Arya, Bran, Jon) spanning the other loop in this case).}
	\label{fig:social_network}
\end{figure}

Simplicial complex is hence a model of the relational structure in the data. Relational structure can be modelled by a graph but graphs only consider pairwise relations between points. In many cases it makes sense to use higher-dimensional connectivity instead modelled with simplices. As a justification consider the example explained in Fig. \ref{fig:social_network}. More fundamental reason is that the simpicial approach views data as spaces spanned by their points and enables the use of powerful mathematical machinery of algebraic topology for the analysis of these spaces, as will be outlined in the following section.

For persistence analysis we define the relation to be a function \(R\) on the data with values in \(\mathbf{R}=[0,\infty)\), i.e. \(R(x,y) \mapsto t \in \mathbf{R}\) for data points \(x\) and \(y\). Concretely we say that \(k+1\) data points \(x_i\) create a $k$-simplex at scale \(t\) if the points satisfy pairwise \(R(x_i,x_j) \le t.\) This construction is called the Vietoris-Rips simplicial complex at scale \(t\). At fixed scale we can then study the connectivity structure. As a standard example, when data is endowed with a distance measure, clustering at some fixed scale corresponds to the 0-dimensional connectivity by only looking at the connected components of the simplicial complex. Simplicial complexes can also contain 1-dimensional connectivity information in the form of loops and holes (see Fig. \ref{fig:social_network}), 2-dimensional information in the form of voids or cavities, etc. These are collectively called topological features. 

Persistence aims to quantify the topological features in a data set and use this information for data analysis. Loop structure might signal about a recurrent dynamics of the phenomenon behind the data. Various dimensional voids can  mark lack of information and connectivity or insufficient data collection. Finding such voids in data sets has aroused interest in different areas of data analysis community, see for example \cite{BigHoles} and references therein. As noted in \cite{BigHoles}, voids can also indicate non-allowed combinations of feature values of data vectors.

One immediate difficulty arises in the simplicial modelling above: what is the appropriate scale of \(R\) to capture the connectivity in various dimensions of an arbitrary set of points? Persistence circumvents this by forming simplicial complexes at all scales \(t \ge 0\) and capturing the evolution of topological features. If a simplex is generated at scale \(t\) it is then present at any subsequent scale and the simplicial complexes are connected by inclusions: \(\cdots \subseteq C_a \subseteq C_b \subseteq C_c \subseteq \cdots\) for \(\dots \le a \le b \le c \le \dots\) The end result of the modelling step is then a mapping called filtration, \((D,R) \times \mathbf{R} \to (C_t,\subseteq_t)_{t \in \mathbf{R}}\), where \((D,R)\) denotes a data set with real-valued relation and \((C_t,\subseteq_t)_{t \in \mathbf{R}}\) denotes an \(\mathbf{R}\)-parameterized sequence of simplicial complexes and inclusions.

\subsection{Algebraic fingerprinting}
Filtration contains all the information about the relations in the data set on various scales. It is therefore very complicated object for infering the global structure of data and simplification is thus necessary. TDA employs tools from mathematical field of algebraic topology, essentially it uses homology of simplicial complexes which transforms the geometric information into algebraic information. We will outline the algorithm for computing homology to illustrate its very implementable nature and to gain intuition on why we are interested in homology in data analysis. For details into homology and its computation see \cite{Rotman,EdelsbrunnerHarer,Mischaikow}. For simplicity we fix the field of coefficients to be \(\mathbf{F}_2\), the field with two elements 0 and 1. Let \(C\) be a simplicial complex and denote by \(C_k\) its set of \(k\)-simplices. Concretely, \(C_0\) consists of the points of the original data set. 

1) Choose an ordering (starting from zero) on \(C_0\) and use it to order elements in any simplex. If \(\{\text{Arya}, \text{Bran}, \text{Jon}\}\) is a 2-simplex in Fig. \ref{fig:social_network}, fix the order in which the points are listed and denote this ordered simplex by \([\text{Arya}, \text{Bran}, \text{Jon}].\) 

2) For natural numbers \(k\) and \(0\leq i\leq k\) and a simplex \(\sigma\) in \(C_k\), define a function \(d_i \colon C_k \to C_{k-1}\) such that \(d_i(\sigma)\) is a simplex in \(C_{k-1}\) formed by removing from \(\sigma\) its \(i\)-th element. The ordering on \(C_0\) was needed to specify the \(i\)-th element in a simplex. For example, \(d_1([\text{Arya}, \text{Bran}, \text{Jon}]) = [\text{Arya}, \text{Jon}].\)
	
3) For any natural number \(k\), let \(\Delta(C)_k\) be the vector space over \(\mathbf{F}_2\) with a base given by all simplices in \(C_k\). An element \(\tau\) in \(\Delta(C)_k\) is then given by a linear combination \(\tau = \sum_{\sigma\in C_k} t_\sigma\sigma, \ t_\sigma \in \mathbf{F}_2.\) The base for \(\Delta(C)_2\) of the simplicial complex in Fig. \ref{fig:social_network} would be [Arya, Bran, Jon] whereas [Arya, Bran]+[Bran, Jon]+[Arya, Jon] would be linear combination of three basis elements in \(\Delta(C)_1\).
	
4) Define \(\partial_k \colon \Delta(C)_k \to \Delta(C)_{k-1}\) to be the linear function assigning to a base element given by a simplex \(\sigma\) in \(C_k\) the linear combination \(\sum_{i=0}^{k}d_i(\sigma)\) of \(k\)-1-simplices. The map \(\partial_k\) is called the boundary operator. Then \(\partial_k([\text{Arya}, \text{Bran}, \text{Jon}]) \allowbreak = [\text{Bran}, \text{Jon}]+[\text{Arya}, \text{Jon}]+[\text{Arya}, \text{Bran}].\) The boundary operator thus formalizes the intuition that \([\text{Bran}, \text{Jon}]+[\text{Arya}, \text{Jon}]+[\text{Arya}, \text{Bran}]\) forms the boundary of \([\text{Arya}, \text{Bran}, \text{Jon}].\) Define \(\Delta(C)_{-1} = 0\) and \(\Delta(C)_k = 0\) for \(k>m\), where 0 denotes the zero vector space.
	
5) The boundary operators connect the various simplices of a simplicial complex together. Computationally the matrices of boundary operators store the global connectivity information in their elements, with coefficient field \(\mathbf{F}_2\) these are just binary matrices. Homology on degree \(k\) of a simplicial complex \(C\) (over coefficients \(\mathbf{F}_2\)) is then defined as a quotient vector space:
\[H_k(C)=	\frac{\text{kernel of }\partial_k\colon  \Delta(C)_k \to \Delta(C)_{k-1}} {\text{image of }\partial_{k+1}\colon  \Delta(C)_{k+1}\to  \Delta(C)_{k}}, \ \ \text{for } k \geq 0.\] 

As noted in step 4) above, some 1-simplices might form the boundary of a 2-simplex. Some 1-simplices on the other hand might form the boundary of an actual hole in the simplicial complex as in Fig. \ref{fig:social_network}. Similarly some \(k\)-1-simplices might form the boundary of a \(k\)-simplex and some might form the boundary of a \(k\)-dimensional hole. By its definition homology quotients out linear combinations of simplices that are boundaries and we are left with those that actually represent linearly independent \(k\)-dimensional holes in the complex. For \(k=0\), \(H_0\) measures the number of linearly independent points that make up boundaries of 1-simplices, effectively the number of connected components.

Homology thus gives us exactly the global connectivity information of the relational structure of data that we seek. The full complexity of a filtration is now simplified by applying homology on degree \(k\). Each simplicial complex is turned into a homology vector space and the inclusion functions are turned into linear maps. The result is an \(\mathbf{R}\)-parameterized sequence of vector spaces and linear maps: \(\cdots \to H_k(C_a) \to H_k(C_b) \to H_k(C_c) \to \cdots.\) We will abbreviate \(H_k(C_a)\) as \(H_{k,a}\). In this parameterized sequence the dimensions of homology vector spaces encode topological information: \(H_{0,t}\) effectively measuring the number of connected components, \(H_{1,t}\) measuring the number of one-dimensional holes and \(H_{k,t}\) those of \(k\)-dimensional voids at scale \(t\). 

This algebraic step gives a mapping \((C_t,\subseteq_t)_{t \in \mathbf{R}} \to (H_{k,t},\to_t)_{t \in \mathbf{R}}.\) The obtained result is not an arbitrary \(\mathbf{R}\)-parameterized vector space. The vector spaces \(H_{k,t}\) are finite dimensional and there are finitely many numbers $0<t_0<\cdots <t_n$ in \(\mathbf{R}\) such that the map $H_{k,a}\to H_{k,b}$ may not be an isomorphism only if $a<t_i\leq b$, for $i$ in \(\{0,\dots,n\}\). These considerations follow from the fact that data sets always contain only finite number of points so topological changes in the relational structure can only occur in discrete steps. Such parameterized vector spaces are called tame \cite{Noise_paper}. An essential result in persistence theory is that any tame \(\mathbf{R}\)-parameterized vector space decomposes into interval indecomposables called bars and the collection of bars in such a decomposition is unique \cite{Carlsson2005}. Bars are enumerated by pairs of numbers $b<d$ in \(\mathbf{R}\). The bar $[b,d)$ at scale $t$ is either a one dimensional vector space, if $b\leq t<d$, and the zero vector space otherwise. The maps between any non-zero vector spaces in a bar are isomorphisms. For a bar $[b,d)$, some topological feature is understood to have appeared in the simplicial complex at filtration value $b$. It is then present in the subsequent simplicial complexes until filtration value $d$. For example, points in the data might connect to create a 1-dimensional loop. This loop persists until at some larger filtration value the points connect further to higher dimensional simplices and the loop vanishes. The bar decomposition can be visualized in a stem plot on a \((b,d-b)\)-coordinate system as shown later in Fig. \ref{fig_activities_density_contour}.

\section{Topological learning}\label{sec:topo_learning}
The actual data analysis step in persistence pipeline is to infer information from the \(\mathbf{R}\)-parameterized sequence of homology vector spaces and linear maps obtained from the map \((D,R) \times \mathbf{R} \to (C_t,\subseteq_t)_{t \in \mathbf{R}} \to (H_{k,t},\to_t)_{t \in \mathbf{R}}\) constructed above. To simplify notation we let \(\mathbf{R}\)-\textbf{Vec} denote the space of tame \(\mathbf{R}\)-parameterized sequences of vector spaces \(V=\cdots \to V_a \to V_b \to V_c \to \cdots.\) Our framework of extracting information from objects in this space is through stabilizing a rank invariant attached to them. Aim of the paper is on the practical data analysis aspects and we only outline the theoretical backgound. For more details we refer to \cite{Noise_paper,RiihimakiWojtek,StableInvariants}.

\subsection{Rank invariant}
The rank, or the dimension, is the fundamental invariant characterizing vector spaces. Similarly we want to assign rank for sequences of vector spaces in \(\mathbf{R}\)-\textbf{Vec}. Let \(V\) be in \(\mathbf{R}\)-\textbf{Vec}. Due to tameness there is a sequence \(0 < t_0 < \cdots < t_k\) in \(\mathbf{R}\) such that $V_a \to V_b$ is not an isomorphism only if $a < t_i \leq b$. Recall that for a linear map \(f \colon X \to Y\) its cokernel is the quotient vector space of \(Y\) by the image of \(f\): \(\text{coker} f = Y / \text{im}f.\) We then define 
\[\beta_0(V) =V_0 \oplus \text{coker}(V_0 \to V_{t_0}) \oplus \text{coker}(V_{t_0} \to V_{t_1}) \cdots\oplus \text{coker}(V_{t_{k-1}} \to V_{t_k}),\]
where \(V_0\) is the homology vector space in \(V\) at filtration value 0. Let us consider what information \(\beta_0(V)\) carries. Since the maps \(V_{t_i} \to V_{t_{i+1}}\) are not isomorphisms the cokernels may not be zero. The quotient by the image removes from the homology vector space \(V_{t_{i+1}}\) the generators, or basis elements, which come from previous non-isomorphic homology vector space. \(\beta_0\) is thus a vector space of the new homology generators that appear in the sequence of homology vector spaces. In the context of filtrations of input data sets, this is a way of keeping track of how topological features created by the relational structure evolve in the simplicial complexes of the filtration. 

For \(V\) in \(\mathbf{R}\)-\textbf{Vec}, its rank is now defined to be a discrete invariant given by the number 
\begin{align*}
	\text{rank}(V)&=\text{dim}(\beta_0(V))=\\
	\text{dim}(V_0) + \text{dim(coker}(V_0 \to & V_{t_0})) + \cdots+ \text{dim(coker}(V_{t_{k-1}} \to V_{t_k})).
\end{align*}

\subsection{Hierarchical stabilization and contour metrics}
The rank defined above is not a stable invariant. Effectively the number \(\text{rank}(V)\) measures the smallest number of homology generators of \(V\). A small perturbation of input data can result in a number of non-essential homology generators. We therefore seek to stabilize the rank invariant to deal with inherent noise in data. Our approach is a general framework for stabilizing discrete invariants.

Let \(T\) be a set of interesting objects and \(I\) the attached invariant. For us \(T\) is of course a collection of \(\mathbf{R}\)-parameterized vector spaces associated to data sets with \(\mathbf{R}\)-valued relation and \(I\) is the rank. The key in converting a discrete invariant into a stable one is to choose a (pseudo)metric \(d\) on \(T\). Once a metric is chosen, we can define an \(\epsilon\)-radius ball around \(X \in T\), \(B(X,\epsilon)= \{Y \ |\ d(X,Y)\leq \epsilon\}\), and look at the function \(\widehat{I}_d(X)\) taking the minimum value of \(I\) on balls around \(X\) with increasing radii \(\epsilon\): 
\[\widehat{I}_d(X)(\epsilon)= \text{min}\{I(Y)\ |\ Y \in B(X,\epsilon)\}.\]

Since we are minimizing the invariant in larger and larger balls around \(X\), the function \(\widehat{I}_d(X)\) is decreasing and piecewise constant, namely a simple function. Due to being a decreasing function with non-negative values, there is some \(t\) such that for all \(s \ge t\) in \(\mathbf{R}\), \(\widehat{I}_d(X)(s) = \widehat{I}_d(X)(t)\). The function \(\widehat{I}_d(X)\) is thus eventually constant with a limit, \(\text{lim}\ \widehat{I}_d(X)\).

The needed metrics in the stabilization can be shown \cite{StableInvariants} to arise from so called contours. Contour is function \(C: \mathbf{R} \times \mathbf{R} \to \mathbf{R}\) satisfying the following inequalities for all \(v,w,\epsilon,\tau\) in \(\mathbf{R}\):
\begin{enumerate}
	\item \(v \leq C(v,\epsilon) \leq C(w,\tau)\), for \(v \leq w\) and \(\epsilon \leq \tau,\)
	\item \(C(C(v,\epsilon),\tau) \leq C(v,\epsilon + \tau)\).
\end{enumerate}

For example, \(C(v,\epsilon) = v + \epsilon\), \(C(v,\epsilon) = v + \epsilon^2\) and \(C(v,\epsilon) =r^\epsilon v\) with a positive number \(r\) are all examples of contours. The contour \(C(v,\epsilon) = v + \epsilon\) is called the standard contour. There is a generic way of producing contours. Let \(f \colon \mathbf{R} \to (0,\infty)\) be a function with strictly positive values which we refer to as density. Then it can be shown that the function \(C(v,\epsilon)\) given by 
\[C(v,\epsilon)=v+\int_{y}^{y+\epsilon}f(x)dx,\] 
where for \(v\) in \(\mathbf{R}\), we have taken the unique \(y\) in \(\mathbf{R}\) such that \(v=\int_{0}^{y}f(x)dx\). For more background on contours we refer to \cite{RiihimakiWojtek}.

It is also shown in \cite{RiihimakiWojtek} how the choice of a contour leads to a pseudometric \(d_C\) in \(\mathbf{R}\)-\textbf{Vec}. The stabilization of the rank invariant with respect to the chosen contour is then defined as  
\begin{equation}\label{eq:stable_rank}
\widehat{\text{rank}}_{C} V(\epsilon)=\text{min}\left\{\text{rank}(W) \ |\ W \in  \mathbf{R}\textbf{-Vec}\text{ and } d_C(V,W)\leq \epsilon \right\}.
\end{equation}
As noted above, the stable rank function \(\widehat{\text{rank}}_{C} V\) is decreasing and piecewise constant and from \(\mathbf{R}\) to \(\mathbf{R}\). 

Our approach does not conceptually rely on the bar decomposition of \(V\) in \(\mathbf{R}\)-\textbf{Vec}. Computation of the decomposition is however standard procedure in persistence analysis with various dedicated implementations \cite{Roadmap} and when the decomposition is given, the stable rank can be computed algorithmically in a very efficient way:
\begin{equation}\label{eq:stable_rank_algorithm}
\widehat{\text{rank}}_C V (\epsilon) = |\{[b_i,d_i)\ |\  C(b_i,\epsilon) < d_i\}|.
\end{equation}
The stable rank of \(V\) at \(\epsilon\) is thus the number of those bars in the decomposition that satisfy the relation between the start and end points given by the contour. In practical computations the limit of \(\widehat{\text{rank}}_C V\) is always zero, or can be set to zero.

By fixing some values of \(\epsilon\) the contour \(C(v,\epsilon)\) reduces to a single variable function and we can plot it. In Fig. \ref{fig_activities_density_contour} this is illustrated with few values of \(\epsilon\) in the stem plot of a bar decomposition. This visualization is helpful in understanding how the contour affects the stable rank in Eq. \ref{eq:stable_rank_algorithm}: the value of stable rank \(\widehat{\text{rank}}_C V (\epsilon)\) at \(\epsilon\) is the number of bars that reach over the function \(C(v,\epsilon)\). If the function \(C(v,\epsilon)\) has lower values it therefore makes bars relatively longer and vice versa with larger values. The contour can thus be seen as controlling pointwise with respect to \(b_i\) the length scale that we use to measure bars.

\subsection{Topological learning with stable ranks}\label{subsec:topo_learning}
The stable rank attached to an input data set is a topological fingerprint of the data. In the actual data analysis task these fingerprints are used in, for example, classifying various data sets. Recall from the construction above that the stable rank is derived by choosing a contour function \(C\) which induces a metric \(d_C\) needed for the stabilization in Eq. \ref{eq:stable_rank}. Each choice of a contour gives a different stable rank capturing different aspects of the data. The learning step in our pipeline is then to choose an appropriate contour for the analysis at hand and we explore this in Section \ref{sec:applications}.

As stable ranks are \(\mathbf{R}\)-valued functions we have various choices of metrics for comparing them. In particular we have standard \(L_p\)-metrics for \(p \ge 1\): 
\[L_p(f,g) =\left( \int_0^\infty|f(t)-g(t)|^p dt \right)^{1/p}.\] 
We can also define interleaving distance between functions \(f\) and \(g\). We first define the set of horizontal shifts of the functions satisfying the indicated inequalities:
\[S =\{\epsilon \in \mathbf{R} \, | \, f(t) \ge g(t+\epsilon) \ \text{and} \ g(t) \ge f(t+\epsilon) \ \text{for all} \ t \in \mathbf{R}\}.\]
The interleaving distance \(d_{\bowtie}\) is then defined as the minimum of those shifts:
\[
d_{\bowtie} (f,g) =
\begin{cases}
\text{inf}(S) &\text{, if $S$ is non-empty,}\\
\infty &\text{, otherwise.}
\end{cases}
\]

In Section \ref{sec:applications} we use these constructions in demonstrating our approach with concrete data analyses. We emphasize that our approach does not rely on any algebraic decomposition of persistence and is thus applicable to multiparameter persistence \cite{CarlssonMultiPers}. The initial theory behind our pipeline was indeed formulated for multiparameter persistence in \cite{Noise_paper} and later specialized for 1-parameter persistence in \cite{RiihimakiWojtek}. In the case of one parameter we obtain the convenient algorithm, Eq. \ref{eq:stable_rank_algorithm}, for computing stable rank.

Traditional view in persistence analysis has been that long bars in the bar decomposition are of importance and smaller bars are noise. This view, however, is challenged by many recent studies showing that smaller features carry important information: study of brain artery trees in \cite{Bendich2016}, functional networks of \cite{Stolz2016}, analysis of protein structure in \cite{XiaWei_protein} and the relation of observed diffraction peaks to small loops in atomic configurations of amorphous silica in \cite{Hiraoka_amorphous}. With our pipeline we can flexibly choose different contours to learn what are in fact the essential features in the data. To produce the bar decompositions we used Ripser software \cite{Ripser}.

\section{Applications}\label{sec:applications}

\subsection{Classifying physical activities}\label{subsec:activities}

We studied PAMAP2 data obtained from \cite{PAMAP} to classify different physical activities. The data consisted of seven persons performing different activities such as walking, cycling or sitting. Test subjects were fitted with three Inertial Measurements Units (IMUs) and a heart rate monitor. Measurements were registered every 0.1 seconds. Each IMU measured 3D acceleration, 3D gyroscopic and 3D magnetometer data. One data set thus consisted 28-dimensional data points indexed by 0.1 second timesteps. 

\begin{figure}[t]
	\begin{subfigure}{0.5\textwidth}
		\centering
		\includegraphics[scale=0.35,clip,trim=5cm 1cm 4cm 0cm]{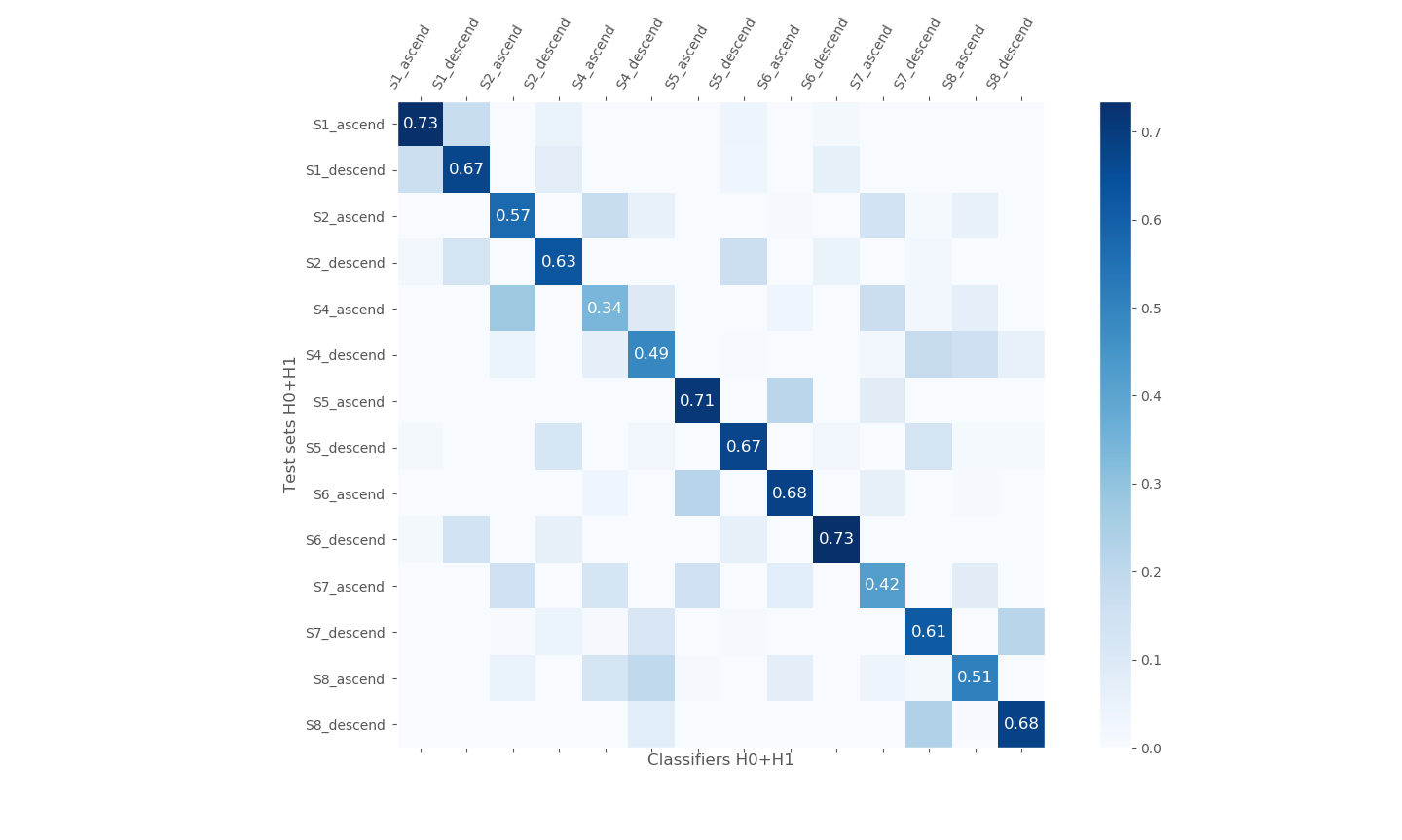}
	\end{subfigure}
	\begin{subfigure}{0.5\textwidth}
		\centering
		\includegraphics[scale=0.35,clip,trim=5cm 1cm 4cm 0cm]{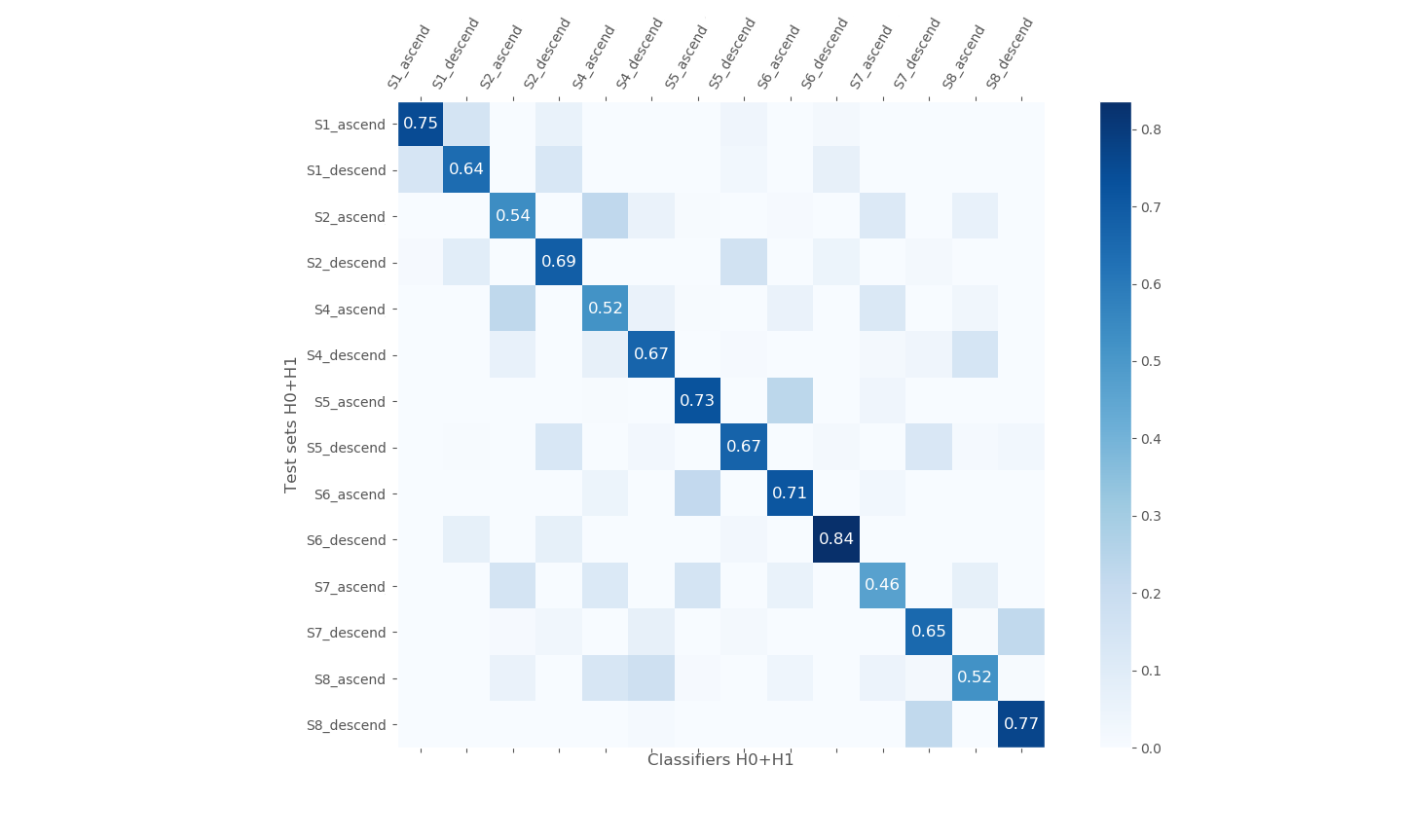}	
	\end{subfigure}
	\caption{Confusion matrices for the classification of ascending and descending stairs activities with standard contour (left) and with contour visualized in Fig. \ref{fig_activities_density_contour}.}
	\label{fig_activities_confusions}
\end{figure}

\begin{figure}[t]
	\centering
	\begin{minipage}{0.99\textwidth}
		\includegraphics[width=0.49\textwidth,clip,trim=3.5cm 1.5cm 1.5cm 2.5cm]{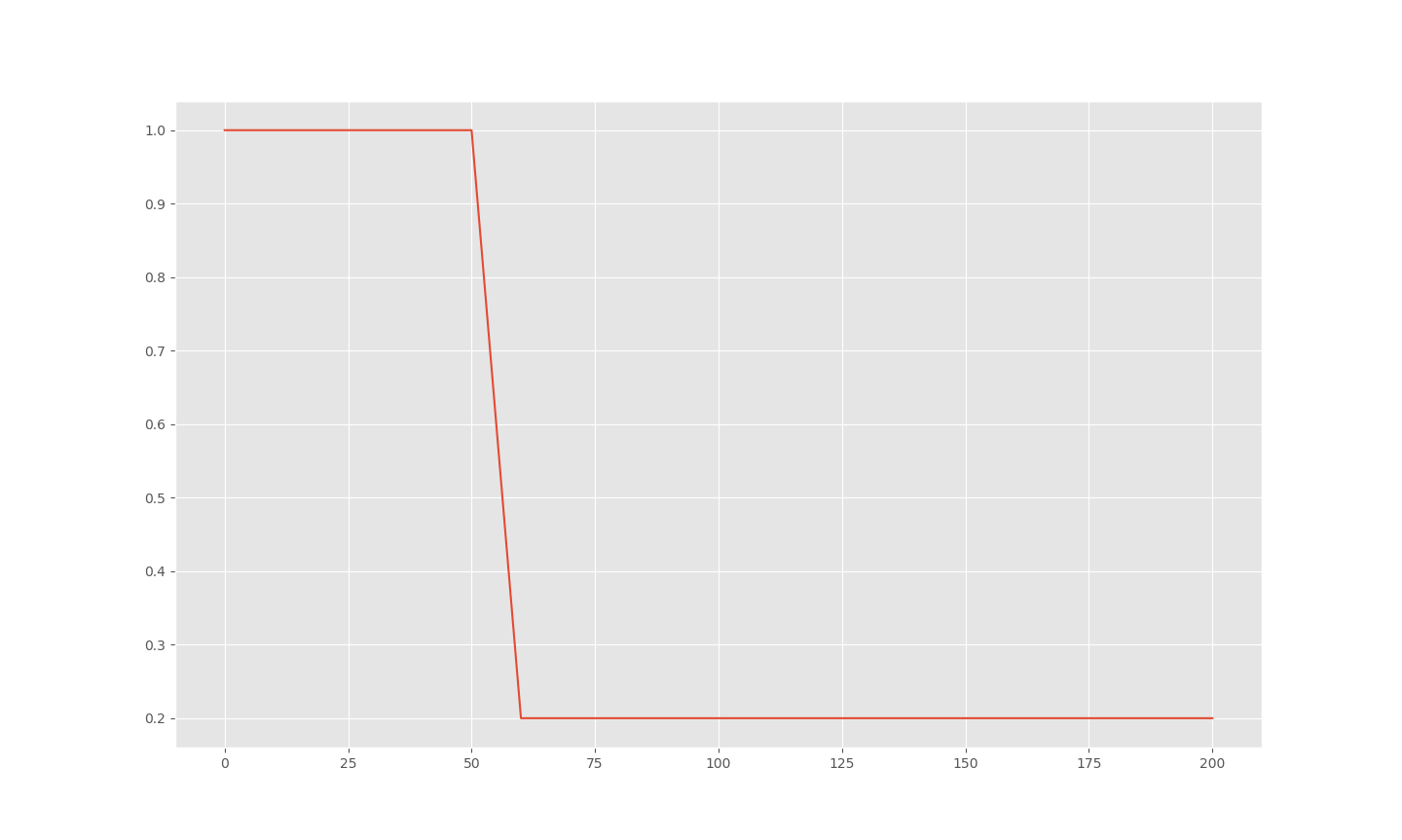}
		\includegraphics[width=0.49\textwidth,clip,trim=3.5cm 1.5cm 1.5cm 2.5cm]{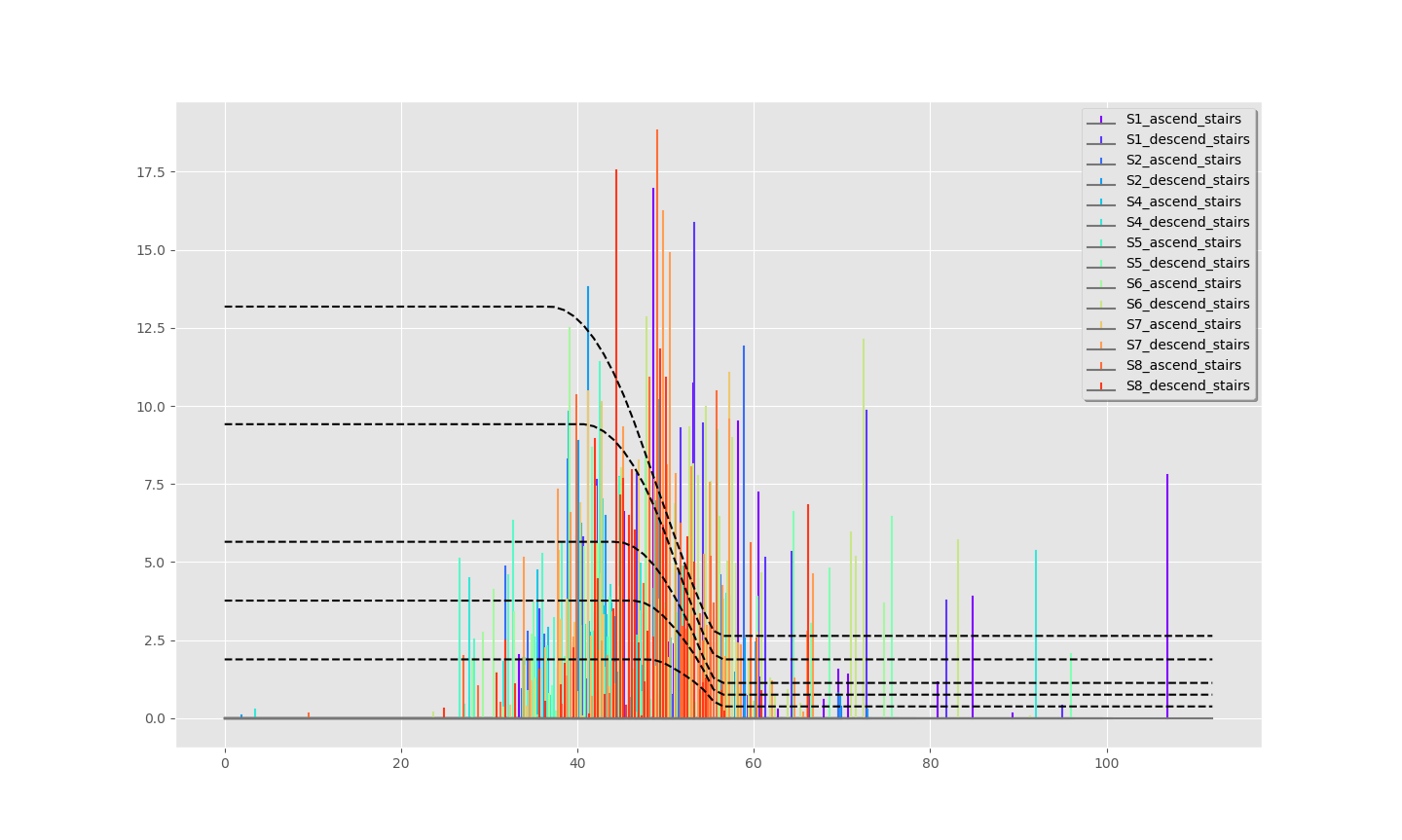}
	\end{minipage}
	\caption{Density function used for $H_1$ stable rank in the activities classification (left) and contour lines for few values of $\epsilon$ (right). Persistence bar stems are shown for single data sets from each (subject,activity) class.}
	\label{fig_activities_density_contour}
\end{figure}

We looked at two activities which from the outset are very similar and expected to be difficult to distinguish: ascending and descending stairs. For the analysis we randomly sampled without replacement 100 points from each data set, repeated 100 times. For each subject we thus obtained 100 resamplings from the activity data and computed their stable ranks with respect to a chosen contour. Out of these we computed the point-wise means of 40 stable ranks in $H_0$ and $H_1$. These means were used as classifiers, denoted by $\hat{P}_{H_0}$ and $\hat{P}_{H_1}$. Altogether we had 14 classifier pairs $(\hat{P}_{H_0},\hat{P}_{H_1})$ corresponding to all (subject, activity) combinations. Remaining 60 stable ranks in $H_0$ and $H_1$ were used as test data and denoted by $T_{H_0}$ and $T_{H_1}$. For a test pair $(T_{H_0}, T_{H_1})$ we found
$$\text{min}(L_1(\hat{P}_{H_0},T_{H_0}) + L_1(\hat{P}_{H_1},T_{H_1}))$$
by computing \(L_1\) distances between the test pair and all classifier pairs. The classification is successful if the minimum is obtained with $\hat{P}_\bullet$ and $T_\bullet$ belonging to the same (subject, activity) class in both \(H_0\) and \(H_1\).

For cross validation we randomly sampled which of the stable ranks constitute classifier and which are test data for the class. Result for 20-fold cross validation is shown in the confusion matrix on the left in Fig. \ref{fig_activities_confusions} for the standard contour. Each cell of the confusion matrix is the number of classifications in the corresponding classifier (columns) and test data (rows) pair relative to the total number of test stable ranks which was 60. Correct classifications are on the diagonal. Overall accuracy (mean over diagonal of the confusion matrix) with standard contour was 60\%. 

We then repeated the above cross validation process but using a different contour in computing $H_1$ stable rank. Contour was obtained from the density function on the left side of Fig. \ref{fig_activities_density_contour}. Contour lines and the bars from persistence computation are visualized on the right side of Fig. \ref{fig_activities_density_contour}. This contour puts more weight on topological features appearing with larger filtration scales. Cross-validation results are shown on the right in Fig. \ref{fig_activities_confusions}. Overall accuracy increased to 65\%. Note particularly increase in the accuracy of subject 4. Also noteworthy is that ascendings mainly get confused with ascendings of different subjects and the same for descendings. These (subject,activity) data thus exhibit different character and changing the contour we could make this difference more pronounced. 

\subsection{Cloud pattern characterization}
We analysed the spatial distribution of shallow cumulus clouds. These clouds form in fair-weather
conditions due to the convective transport of heat and moisture in the atmosphere. Convection is a classic example of a pattern-forming system
\cite{Mizushima1994,Cerisier1996}. Cloud formation is known to be influenced by diverse physical processes across spatial scales ranging from
molecular sizes to kilometers. Such spatial scales and all their physical variables cannot be explicitly resolved in numerical climate models,
which calls for the development of cloud parametrization schemes.
Moreover, the spatial distribution of clouds influences their formation processes. It is therefore important to include this distribution
in parametrization schemes. This problem has been studied from different perspectives, notably the influence of land
surface conditions on cloud formation~\cite{Rieck2014}. Here we describe an approach based on persistence and the use of stable ranks as
descriptors of the spatial distribution of clouds. See \cite{LiconRiihimakivanLaar} for further results and references. 

\begin{figure}[t]
	\centering
	\includegraphics[width=\textwidth]{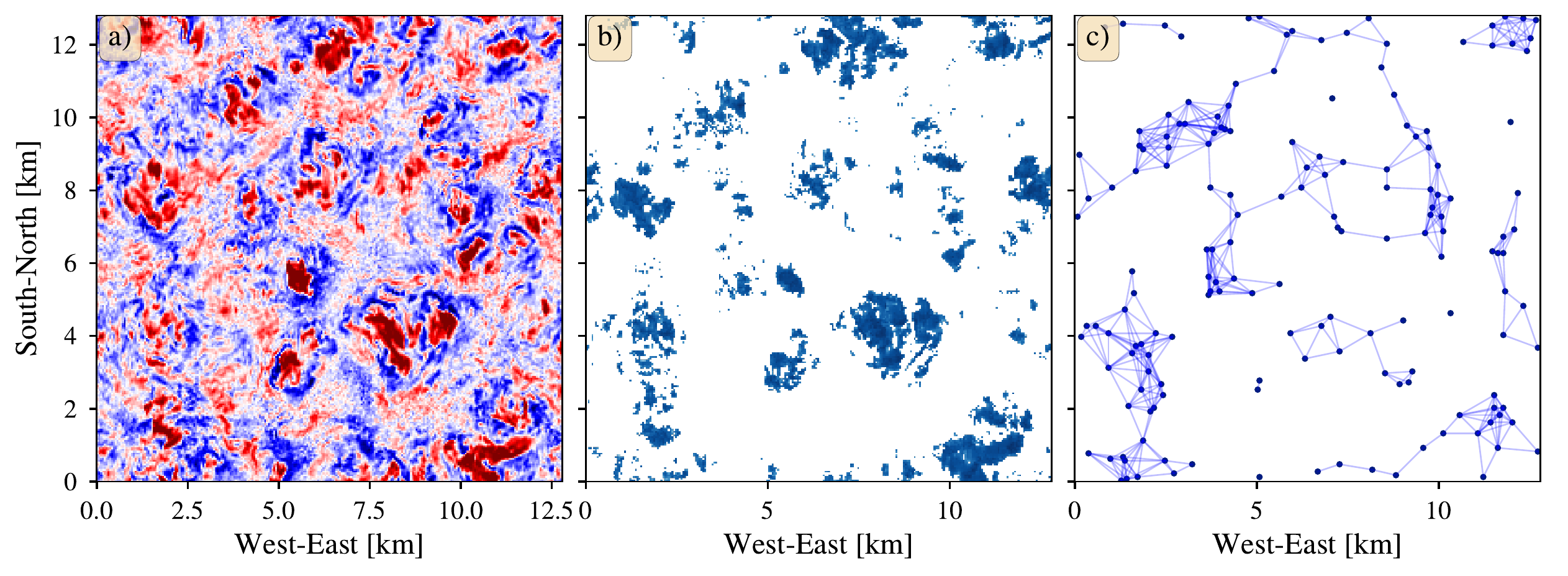}
	\caption{a) Values of the vertical wind velocity $w$ for a two-dimensional horizontal slice at an altitude of
        \SI{1.8}{\kilo\metre}. This corresponds to cloud base height (red -- $w > 0$; blue -- $w < 0$).
        b) Column liquid water content $ql$ (i.\,e. the maximum liquid water value in the vertical direction).
        c) Point representation of the cloud field by the local maxima of $ql$
        (only connected components formed by at least 3 cells are considered),
        and 1-simplices of the Vietoris-Rips filtration using the distance relation between the points, at a distance scale of \SI{1.5}{\kilo\metre}.}
	\label{fig:cloud-fields}
\end{figure}

The data was produced by the Dutch Atmospheric Large-Eddy Simulation model and covered the time
period between 09:00h and 18:00h during one day, saved for analysis at 15 minute intervals, with model setup similar to 
that in~\cite{Neggers2012}.
We simulated 10 days with different initial conditions. The data consists of large amount of physical information from which cloud fields can be extracted. The spatial simulation domain in \(x,y,z\) coordinates is \(12.8 \times 12.8 \times 5\) in kilometers with horizontal resolution of 50 meters and vertical resolution of 40 meters. The computation domain thus consists of cells. A homogeneous land surface is prescribed and the lateral boundaries are periodic. The 3D cloud field from the simulation domain was then flattened in the \(z\)-direction onto a 2D plane by
taking the maximum liquid water content, $ql$, values in the vertical direction. The resulting cloud fields are then as visualized in Fig.~\ref{fig:cloud-fields}(b).

An important issue in the study of cloud formation is the quantification of spatial organization, or lack thereof, in a given cloud field. While methods to study spatial distributions exist in the statistical literature for objects which can be idealized as points,
it is harder to work with objects that possess a spatial extent (i.e. area or volume), as clouds do. This leads to the necessity of computing a point representation for a cloud before being able to assess the spatial distribution of the cloud field. Here we consider three
different representations: assigning to each cloud its geometric centroid, its point with maximum $ql$ value, and a set of its 
points chosen at random.

A common metric in the assessment of spatial organization is the \(I_\text{org}\) index~\cite{Tompkins2017}, defined as follows. For a two-dimensional cloud field, such as the one shown in Fig. \ref{fig:cloud-fields}(b), index the connected components (the individual clouds) as $c_i$, and compute their geometric centroids, $\bar c_i$. We are interested in how the spatial distribution of the $\bar c_i$ compares to what we would expect under complete spatial randomness (CSR), that is, if the centroids represent a realization of a homogeneous Poisson point process. To that end, we consider the nearest-neighbor distances $d_i$, which are
defined as $d_i = \text{min}\{d(\bar c_i, x) \ | \ x \in \bar{\mathcal C} \setminus \{ \bar c_i \}\}$, where $\bar{\mathcal C}$ represents the set of all centroids. The cumulative distribution function (CDF) of the $d_i$ is
\begin{equation*}
    G_{d_i}(r) = P[d_i \leq r],
\end{equation*}
which in the case of a Poisson point process has the analytic expression
\begin{equation*}
    G_{CSR}(r) = 1 - \exp{(-\lambda \, \pi \, r^2)},
\end{equation*}
where $\lambda$ is the Poisson intensity parameter. The value of \(I_\text{org}\) is then defined to be the area under the graph $(G_{CSR}(r), \hat G(r))$, where
\begin{equation*}
    \hat G(r) = \frac{\# \{ \bar c_i \in \bar {\mathcal C} \mid d_i \le r \} }{\# \{ \bar c_i \in \bar{\mathcal C} \} }
\end{equation*}
is the empirical estimator of $G(r)$. If $\hat G$ matches well with $G_{CSR}$, the value of \(I_\text{org}\) will be close to \num{0.5}. A value larger than this suggests spatial clustering, while a smaller one suggests dispersion or regularity.

\begin{figure}[t]
\centering
\includegraphics[width=1\linewidth]{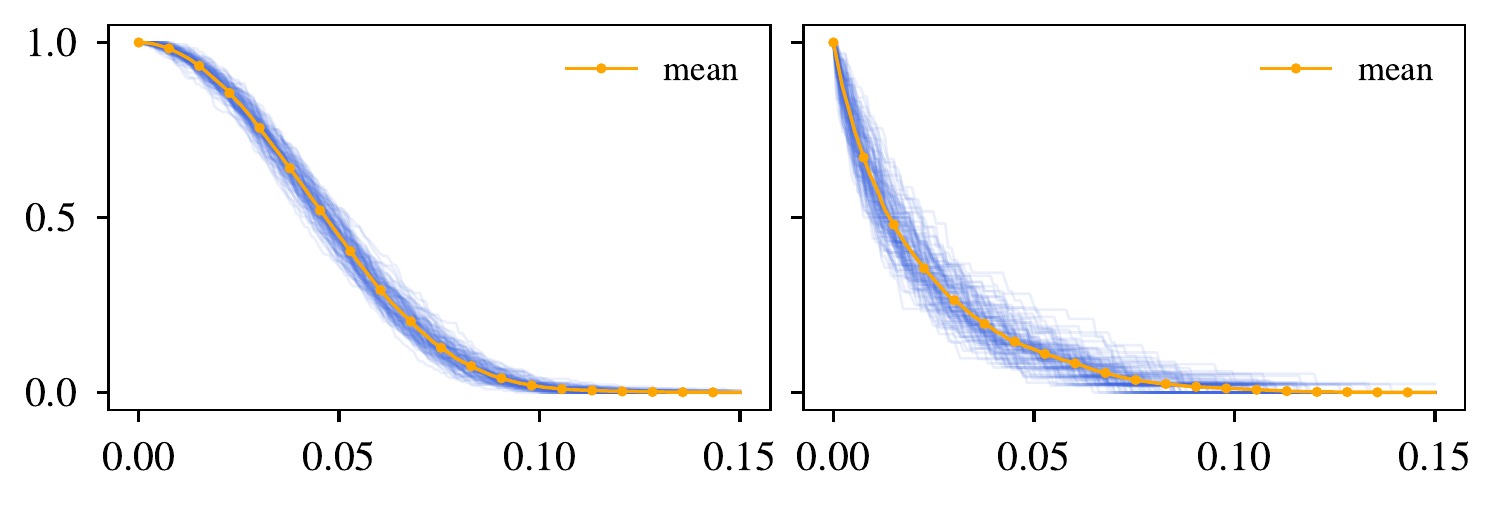}
\caption{Stable rank functions obtained from 100 realizations of a homogeneous Poisson point process with $\lambda = 100$. Left: $S_0^*$. Right: $S_1^*$. }
\label{fig:poisson-sr-example}
\end{figure}

Let $S^*_i$ denote the stable rank of \(H_i\) with respect to the standard contour (Eq.~\ref{eq:stable_rank_algorithm}), normalized by its value at \num{0}. If we define the function $G_{PH}^i(r) = 1 - S_i^*(r)$, we note that it increases monotonically towards \num{1}. In fact, since the normalized stable rank at \(r\) is an indication of the relative amount of homological features that persist beyond \(r\), the function $G_{PH}^i(r)$ can be understood as the empirical CDF of homological persistence.

For $n$ realizations of a Poisson point process with intensity parameter $\lambda$, we find that their normalized stable ranks $S^*_i$, and therefore also $G_{PH}^i$, oscillate within a narrow band 
(see Fig. \ref{fig:poisson-sr-example}). At this point we do not have an analytic expression for the stable rank functions obtained from a Poisson point process, but we can define persistent homology analogues to the \(I_\text{org}\) index via a Monte Carlo procedure by taking the area under the curves defined by $(G_{PH,CSR}^{i}(r), G_{PH}^i(r))$. In the case of a point process in the plane we would then get two values \(I_{PH,0}\) and \(I_{PH,1}\). We define the index as their arithmetic mean,
\begin{equation}\label{eq:ph-index}
    I_{PH} = \frac{I_{PH,0} + I_{PH,1}}{2}.
\end{equation}

\begin{figure}[t]
\centering
\includegraphics[width=1\linewidth]{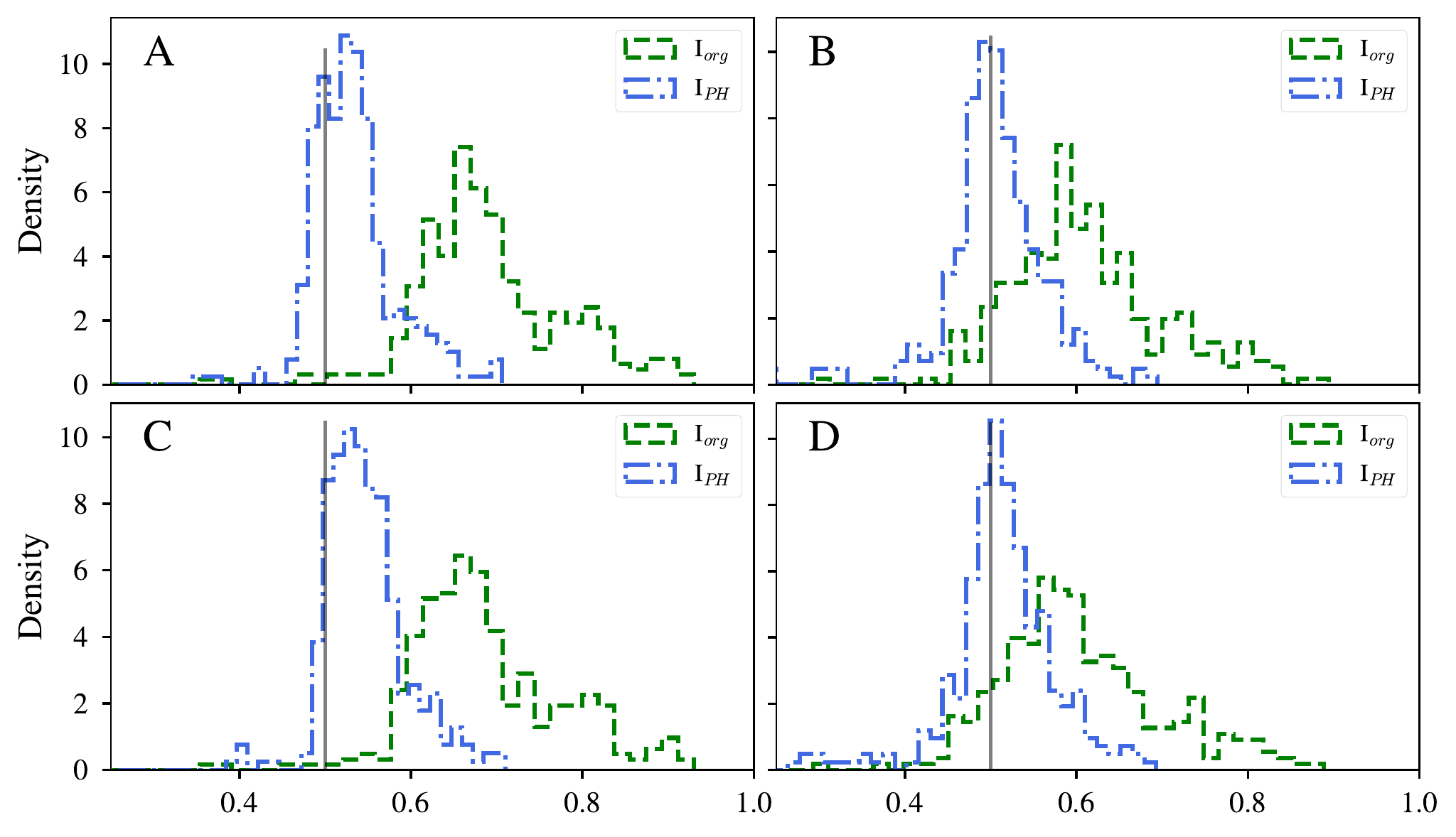}
\caption{Density histograms of the \(I_\text{org}\) index and \(I_{PH}\) (Eq. \ref{eq:ph-index}) for 360 distinct cloud fields. \textbf{A}: $ql$ max, \textbf{B}: $ql$ max removing cloud structures with size smaller than 3 cells, \textbf{C}: Geometric centroids, \textbf{D}: Geometric centroids removing cloud structures with size smaller than 3 cells.}
\label{fig:hist-index-values}
\end{figure}

We tested the performance of the index \(I_{PH}\) defined above, and compared it to the corresponding values
of \(I_\text{org}\) in the dataset consisting of 360 distinct cloud fields (36 per simulation day). The values of both
indices are shown in Fig.~\ref{fig:hist-index-values}. Each panel shows the \num{360} values of each index for all
cloud fields, computed using 4 different point representations. Panel A shows the values obtained from assigning
to each connected component its point with maximum $ql$ value (local maxima); panel B shows the indices obtained
when using the local maxima but only of those components with size at least 3 grid cells (all smaller
components are ignored). Panel C shows the results of using the geometric centroid of each connected component.
Finally, for panel D the geometric centroids were used after discarding the smaller components. These small
components can be attributed to numerical imprecision in the underlying model, and hence are not physically
meaningful.

As discussed above, if these indices have a value close to \num{0.5}, it would indicate that the point
process that they are evaluated on is close to complete spatial randomness, or a Poisson point process.
In the simulations used here, we have cause to expect spatially random behavior: the domain size is too
small to allow for deep convection and spatial organization to happen. Moreover, the lack of land surface
features or patterns means there are no forcings at different spatial scales. Thus the spatial
distribution of physical variables is dominated by the characteristic patterns present in atmospheric
turbulence, itself an essentially random process. The values of the persistent homology index \(I_{PH}\) strongly support this hypothesis, while \(I_\text{org}\) exhibits values in general larger than \num{0.5}. This can be attributed to the fact that it is based on nearest-neighbor distances only, whereas the stable rank functions reflect the spatial relationships of the points throughout all spatial scales. This is confirmed by the fact that removing the smaller structures in the fields (those less than 3 grid cells in size) brings the values of \(I_\text{org}\) closer to \num{0.5} on average, whereas the average
for \(I_{PH}\) is barely affected. This highlights the fact that, by virtue of using all the spatial information available, the persistent homology based method is inherently more robust than any nearest-neighbor method.

\begin{figure}[t]
	\begin{subfigure}{0.5\textwidth}
		\centering
		\includegraphics[scale=0.25,clip,trim=2.7cm 1.5cm 3cm 2.5cm]{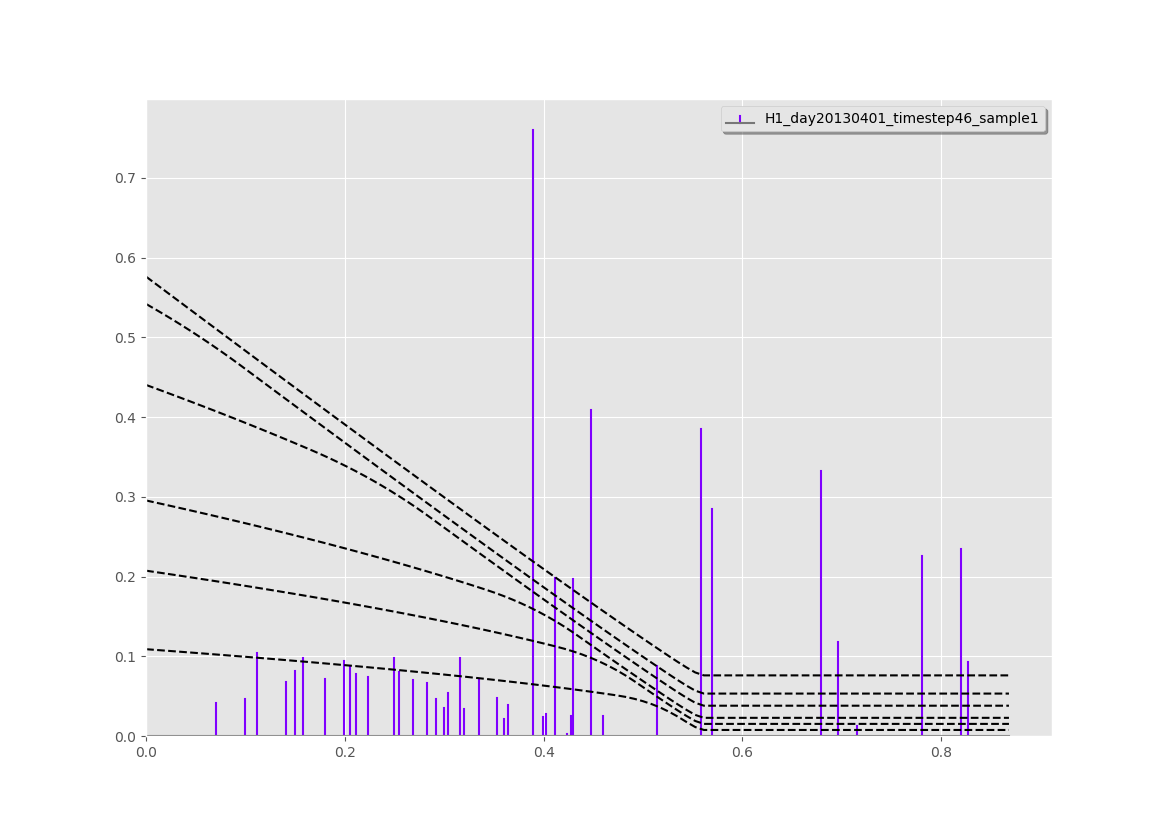}
	\end{subfigure}
	\begin{subfigure}{0.5\textwidth}
		\centering
		\includegraphics[scale=0.25,clip,trim=2.7cm 1.5cm 3cm 2.5cm]{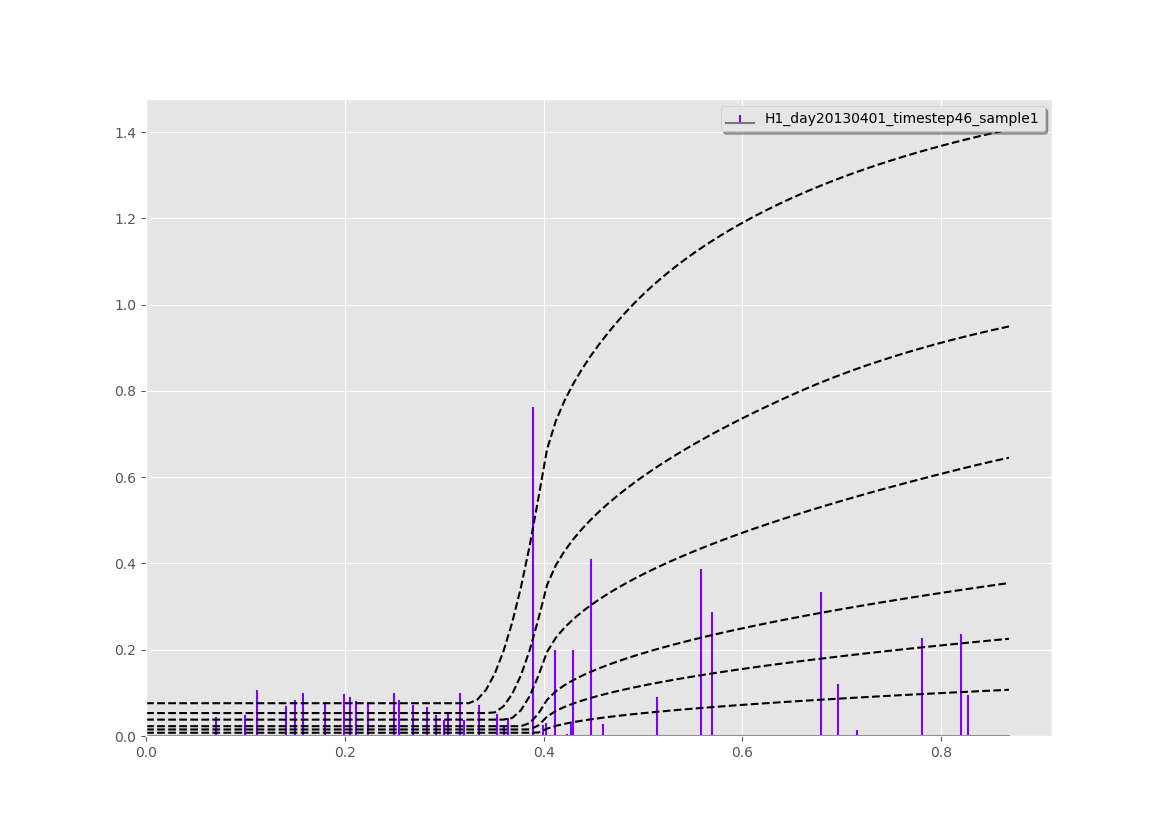}		
	\end{subfigure}
	\caption{Contour 1 (left) and contour 2 (right) used in the analysis of cloud fields. Stem plot is from one sampling of
		a cloud field at one time step.}
	\label{fig_cloud_data_contours}
\end{figure} 

This result has been arrived at by using the standard contour only, which implies that spatial randomness in these cloud
fields is obtained when all spatial scales present in the data are given the same weight. It is possible to obtain different
morphological classifications of the same fields by using alternative contours, which emphasize spatial features differently at varying
scales, as presented with the classification in Section \ref{subsec:activities}. We used standard contour and contours visualized in Fig.~\ref{fig_cloud_data_contours}. These contours are referred to as contour 1, denoted \(C_1\), and contour 2, denoted \(C_2\). 

\begin{figure}[!t]
	\centering
	\includegraphics[width=1\linewidth]{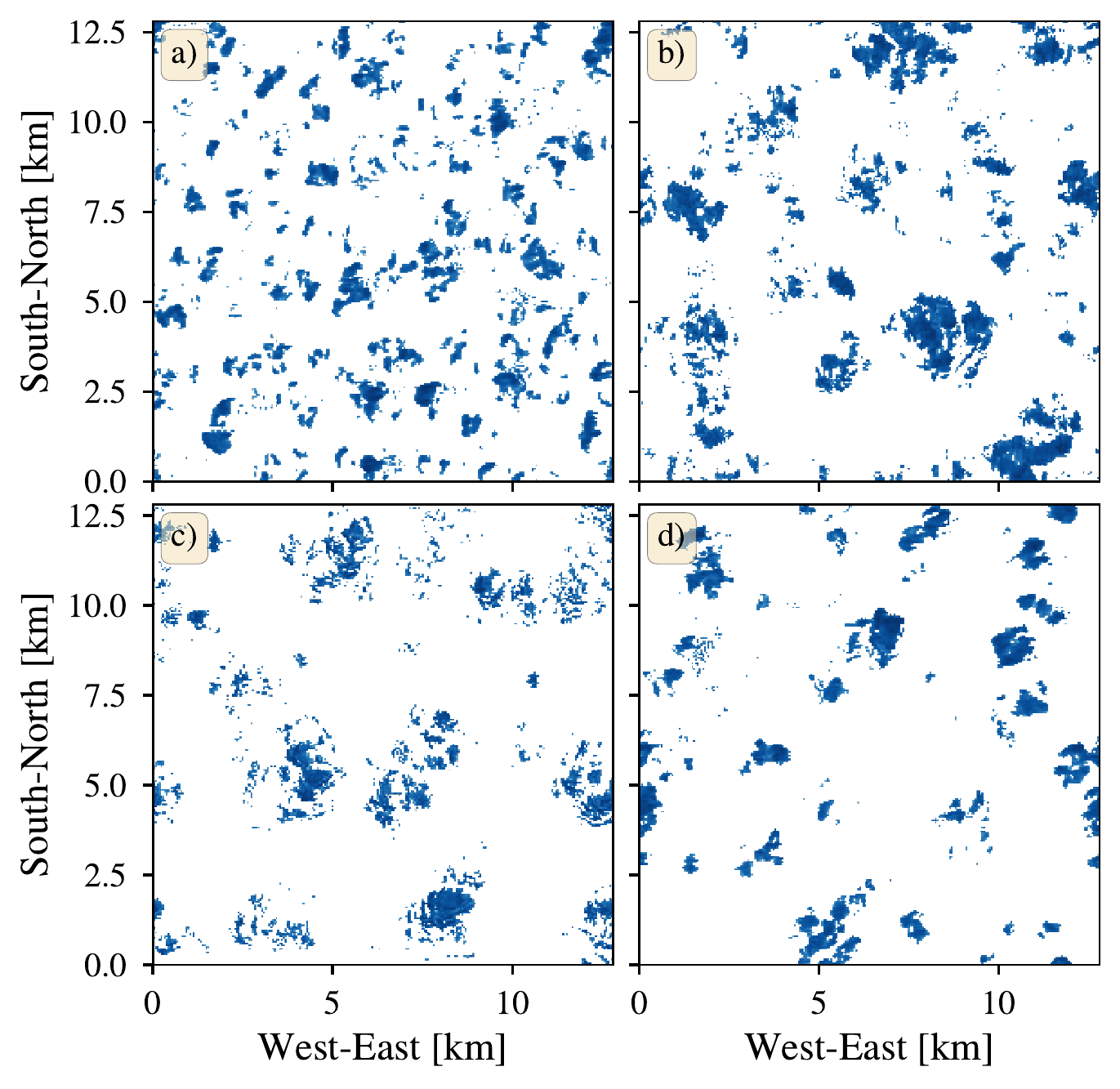}
	\caption{Cloud fields which are classified into different clusters, according to the methodology described in the text.
		We use the $H_1$ stable ranks and the interleaving metric to compute the distances between them.
		a) and b) are classified using contour $C_1$, and have \(I_\text{org}\) values of \num{0.45} and \num{0.53} respectively.
		Cloud cover is similar at \num{14}\% for both.
		c) and d) are classified with $C_2$, and have \(I_\text{org}\) values of \num{0.65} and \num{0.63} respectively, and cloud cover
		for both is \num{9.2}\%.}
	\label{fig:cloud-clusters-example}
\end{figure}

To reduce the effect of sampling, 
10 random samples were drawn from each of the 360 cloud fields, with sample rate 5\% of cloud size. To each cloud field
we assign the mean stable rank of these 10 samples. Stable ranks were computed in \(H_1\) with respect to standard contour, contour 1 and contour 2 and normalized to give $S^*_1$ function as explained above. After removing those cloud fields without \(H_1\) features, we have
254 normalized stable ranks $S^*_1$ for each class of contours. Distance matrices using interleaving, \(L_1\)- and \(L_2\)-metrics (see Section \ref{subsec:topo_learning}) were then computed for the three different classes of stable ranks. Dendrograms from the distance matrices were visually analyzed to decide on a number of clusters of stable ranks. From these computations the interleaving distance gave the clearest clustering results. With respect to contours, \(C_1\) and \(C_2\) gave better clustering than standard contour.

An example of diverging morphological characteristics educed from the \(C_{1,2}\) clustering schemes is shown in
Fig.~\ref{fig:cloud-clusters-example}:
(a) and (b) are representatives of two different clusters obtained by using contour $C_1$, while (c) and (d) stem from
clusters in the $C_2$ classification. As expected from the definition of the contours, the classifications they induce are influenced by different spatial scales. Namely, despite the fact that cloud fields a) and b) have identical cloud cover, and their \(I_\text{org}\) values are very similar, the large-scale
distribution of the individual clouds is significantly different for both. In similar fashion, both c) and d) are indistinguishable in terms of cloud cover and \(I_\text{org}\), yet are distinguished by the spatial pattern of smaller structures, even if the large-scale distribution is similar in both.

This study of cloud fields shows that the use of stable rank functions as descriptors for spatial distributions can reveal morphological properties which other methods cannot. Crucially, the possibility of changing the contour enriches the scope for determining such properties. Future investigation in this direction will address questions such as: what the optimal contour is for a given problem, what these methods can reveal about the temporal evolution of cloud formation, and how the homological properties thus discovered can be related to different physical variables in the system. From general data analysis point of view, particularly the optimization of contours is crucial for making our pipeline a full scale machine learning approach.

\section*{Acknowledgments}
We gratefully acknowledge Roel Neggers for providing the DALES simulation data.
JLS acknowledges support by the DFG-funded transregional research collaborative TR32 on Patterns in Soil--Vegetation--Atmosphere Systems.

%
%
\bibliographystyle{splncs04}
%

\end{document}